\documentclass[10pt, a4paper]{article}

\usepackage[final]{lrec2026} 

\usepackage[most]{tcolorbox}
\usepackage{graphicx}
\usepackage{booktabs}
\usepackage{arydshln}
\usepackage{amsmath}
\usepackage{colortbl,array,xcolor}
\usepackage{tabularray}
\usepackage{multirow}
\usepackage{amssymb}

\usepackage{CJKutf8}
\usepackage[utf8]{inputenc}
\usepackage[T1]{fontenc}
\newcommand{\ja}[1]{\begin{CJK}{UTF8}{ipxm}#1\end{CJK}}

\definecolor{rowgray}{gray}{0.5}
\newtcolorbox{oogiriexample}[1][]{
  colback=gray!3, colframe=gray!60,
  boxrule=0.4pt, arc=1pt, left=6pt, right=6pt, top=2pt, bottom=2pt,
  sharp corners, before skip=3pt, after skip=3pt, #1
}
\newcommand{\bb}[1]{\textbf{#1}}
\newcommand{\prompt}[1]{{\small \textbf{Prompt:} #1}}
\newcommand{\response}[1]{{\small \textbf{Response:} #1}}
\newcommand{\datasetName}{\textsc{Oogiri-Corpus}}
\newcommand{\benchmarkName}{\textsc{Oogiri-Master}}

\title{\textsc{Oogiri-Master}: Benchmarking Humor Understanding via Oogiri}

\name{Soichiro Murakami$^{1}$, Hidetaka Kamigaito$^{1,2}$, Hiroya Takamura$^{3}$, Manabu Okumura$^{3}$}

\address{$^{1}$CyberAgent, $^{2}$Nara Institute of Science and Technology, $^{3}$Institute of Science Tokyo \\
         murakami\_soichiro@cyberagent.co.jp, kamigaito.h@is.naist.jp, \{takamura,oku\}@pi.titech.ac.jp\\
        }

\abstract{
Humor is a salient testbed for human-like creative thinking in large language models (LLMs).
We study humor using the Japanese creative response game Oogiri, in which participants produce witty responses to a given prompt, and ask the following research question: \textit{What makes such responses funny to humans?}
Previous work has offered only limited reliable means to answer this question. Existing datasets contain few candidate responses per prompt, expose popularity signals during ratings, and lack objective and comparable metrics for funniness. 
Thus, we introduce \benchmarkName~and \datasetName, which are a benchmark and dataset designed to enable rigorous evaluation of humor understanding in LLMs.
Each prompt is paired with approximately 100 diverse candidate responses, and funniness is rated independently by approximately 100 human judges without access to others’ ratings, reducing popularity bias and enabling robust aggregation.
Using \datasetName, we conduct a quantitative analysis of the linguistic factors associated with funniness, such as text length, ambiguity, and incongruity resolution, and derive objective metrics for predicting human judgments.
Subsequently, we benchmark a range of LLMs and human baselines in \benchmarkName, demonstrating that state-of-the-art models approach human performance and that insight-augmented prompting improves the model performance.
Our results provide a principled basis for evaluating and advancing humor understanding in LLMs.
\\ \newline 
}

\begin{document}

\maketitleabstract

\section{Introduction}
\label{sec:introduction}
\looseness=-1
Endowing large language models (LLMs) with human-like creative thinking capabilities is a major challenge that extends beyond problem-solving abilities.
Humor understanding is one of such key capabilities. 
Understanding and generating humor as humans require more than pattern matching; they necessitate creative reasoning that incorporates context and cultural nuances to produce witty and unexpected responses \cite{Loakman2025INLG}.
This study addresses humor as an instance of creative thinking in LLMs by focusing on the specific case of \textit{Oogiri}~(\ja{大喜利}). 
\textit{Oogiri} is a Japanese creative response game that involves improvising humorous responses to a given prompt, as shown in Figure~\ref{fig:oogiri-examples}, making it an ideal testbed for creativity and wit.
This raises the central question: \textit{What exactly makes Oogiri responses funny to humans?}
The starting point of our study is to answer this question. 
Few studies have aimed to capture the human perception of funniness using objective metrics and to analyze its components quantitatively. 
This absence poses a significant barrier to the evaluation of humor understanding in LLMs.

\looseness=-1
We address two key challenges in evaluating the humor understanding of LLMs.
First, the constituent elements of a funny response remain insufficiently understood.
Humor is a subjective construct arising from a complex interplay of factors such as the violation of expectations and resonance. However, an objective, quantitative metric does not exist for measuring funniness itself.
Consequently, we lack a principled basis for explaining why an Oogiri-style response is funny, which hinders the systematic improvement of LLM humor understanding.
The second challenge is the low reliability of existing datasets for such analysis.
For example, the Oogiri-GO dataset \cite{Zhong_2024_CVPR} was collected from Bokete,\footnote{\url{https://bokete.jp/}} a caption-contest platform on which users upvote funny responses to prompts.
Although this social-voting signal is useful at this scale, it introduces two methodological limitations.
First, the fairness of the evaluation process is not guaranteed: making the popularity of each response visible to other raters may introduce popularity bias and compromise objectivity.
Second, the dataset exhibits structural bias. 
With only approximately eight candidate responses per prompt, on average, raters are likely to select \textit{a relatively better} option rather than an intrinsically humorous one.

\begin{figure}[t]
    \centering
    \begin{oogiriexample}
      \prompt{Worst commit message ever.}\\
      \response{``It works on my machine.''}
    \end{oogiriexample}
\caption{Oogiri prompt--response example.}
\label{fig:oogiri-examples}
\end{figure}

\looseness=-1
Therefore, in this study, we propose \benchmarkName, a benchmark 
that evaluates the humor understanding of LLMs using the Oogiri task.
Specifically, we address the two challenges outlined above by constructing a novel dataset and conducting a quantitative analysis of the funniness components, with which we assess the current capabilities and pave the way for improvements.
First, we construct \datasetName, a dataset that ensures reliability and objectivity.\footnote{We distinguish the dataset, \datasetName, which underpins our analyses, from the benchmark, \benchmarkName, which builds on it to evaluate LLMs.}
On average, each prompt is paired with approximately 100 diverse candidate responses that are rated for funniness by approximately 100 human judges in an independent setting in which they cannot see others’ ratings.
This design mitigates the issues of fairness and data bias observed in existing datasets.
Second, using this dataset, we quantitatively analyze the linguistic features that constitute funniness.
We identify common lexical and structural patterns in high-rated responses, transforming the ambiguous notion of funniness into measurable, objective metrics.
This enables explanations of \textit{why a response is funny} based on data-driven evidence, rather than subjective intuition.
Finally, we present the multifaceted benchmark results on \benchmarkName.
We benchmark humans and various LLMs to clarify the current state of the art in the humor understanding of LLMs.

The contributions of this study can be summarized as follows:\footnote{The dataset and the benchmark will be provided under the CC BY-NC-SA 4.0 license.}
First, we constructed and release a large-scale reliable dataset, \datasetName, which serves as a novel foundation for evaluating humor understanding in LLMs.
Second, through quantitative analysis of this dataset, we identified the constituent components of funniness, demonstrating that features such as response length, perspective shift, and ambiguity are strongly correlated with high-rated responses. 
Third, we propose a novel benchmark, \benchmarkName, and experimentally demonstrated that 
(1) state-of-the-art LLMs such as GPT-5 show performance approaching human performance; 
(2) our analytical insights into the constituent components of humor can contribute to performance improvements in humor judgment; 
(3) instructing LLMs to leverage these insights only when uncertain improves their performance; and simultaneously,
(4) continued pretraining on the target-language corpus enhances the humor understanding abilities of LLMs.


\section{Related Work}
\label{sec:related_work}
\paragraph{Background on Computational Humor}
\looseness=-1
Computational humor is a relatively new area, and humor understanding/generation remains a challenging problem in natural language processing~\cite{Loakman2025INLG}. 
One obstacle is defining ``humor'' appropriately.
Consequently, many studies have narrowed the scope to specific forms (e.g., puns, Oogiri, satire) to make the problem tractable~\cite{amin-burghardt-2020-survey}.
Among these, pun generation has a particularly long history and is a central task~\cite{ritchie-2005-computational,yu-etal-2018-neural,luo-etal-2019-pun}

\paragraph{Oogiri as a Testbed for Humor Understanding}
We target Oogiri as our testbed for humor understanding.
Oogiri is a creative response game in which one provides a witty response to a prompt.
Although the most common setup is a text-to-text format in which a textual prompt is paired with a textual response, modal variants exist (e.g., image-to-text one-liners; image\&text-to-text fill-in-the-blank)~\cite{Zhong_2024_CVPR}.
These formats resemble \textit{memes} \cite{Sharma_Agarwal_Suresh_Nakov_Akhtar_Chakraborty_2023,nguyen-ng-2024-computational}; we regard memes as a multimodal variant of Oogiri.
However, we focus on text-to-text Oogiri for two reasons.
First, abundant web resources exist.
Oogiri is widely popular in TV programs and social media, and large platforms such as Bokete and Oogiri Sogo
host substantial data.
Because analyzing humor components requires diverse and numerous samples, Oogiri is suitable from a data perspective.
Second, the text-to-text format is unimodal, making semantic understanding more straightforward than with multimodal variants.

\paragraph{Existing Oogiri Datasets and Their Limitations}
Although progress has been hampered by limited datasets, interest has recently increased with the advent of LLMs and the concomitant need for evaluation resources.
Oogiri-specific datasets remain relatively scarce; adjacent resources include English caption datasets collected from the New Yorker Caption Contest \cite{hessel-etal-2023-androids} and various meme datasets~\cite{liu-etal-2022-figmemes,hwang-shwartz-2023-memecap,hossain-etal-2022-memosen}.
Oogiri-GO, which was built using Bokete and social media, is a representative Oogiri dataset. 
However, it faces two issues:
(1) fairness concerns: Voter interfaces display others’ popularity, inviting conformity and potentially compromising objectivity.
(2) structural bias: Many prompts have few candidate responses (approximately eight on average); hence, raters may select responses that are merely ``less bad,'' rather than intrinsically funny.
In this study, we construct a novel Oogiri dataset, \datasetName, which addresses these issues and serves as a foundation for evaluating LLM humor understanding, thereby improving reliability.

\paragraph{Quantitative Analyses of Humor Components}
Although studies have been conducted on generation, understanding, and explanation in computational humor \cite{amin-burghardt-2020-survey,Loakman2025INLG}, quantitative analyses of the constituent components of ``funniness'' remain underexplored.
To fill this gap, using \datasetName, we analyze how diverse linguistic features, such as perspective shift, ambiguity, harmlessness, surprisal, sentence length, and part-of-speech (POS) ratios, relate to humor, with the aim of identifying objective, quantitative indicators.
Furthermore, using our benchmark experiments, we outline how these insights can improve LLM humor understanding.

\section{Dataset Construction}
\label{sec:dataset_construction}
\looseness=-1
Motivated by the second challenge mentioned in \S\ref{sec:introduction}, we present \datasetName~and provide details on its construction process and descriptive statistics.
We collected data from a public Japanese Oogiri competition platform, Oogiri Sogo\footnote{\url{https://chinsukoustudy.com/}}.
On this platform, each prompt proceeds through an answer phase, a voting phase, and a final leaderboard announcement.
During the answer phase, users submit responses within a fixed time window (e.g., 12 h).
This phase then transitions to the voting phase, in which users vote for the responses that they find funny among all submissions.
Unlike other platforms (e.g., Bokete), vote counts are not displayed during the voting phase, which helps to mitigate popularity bias and supports fairer evaluation.
Finally, the platform announces a leaderboard based on the total votes.

\looseness=-1
Dataset construction comprised two steps: web crawling\footnote{The site explicitly permits web crawling.} and quality filtering.
First, we collected 2,165 prompts from the platform.\footnote{Prompt IDs 87--2254 were available when accessed.}
Each prompt is associated with many responses, and each response has a vote count indicating its perceived funniness.
We applied vote-based filtering to ensure reliability: we excluded prompts for which the total number of votes 
was fewer than 100.
This threshold reduces the variance owing to rater subjectivity and chance when the vote pool is small.
In total, 908 prompts remained.
We refer to this 908-prompt dataset as \datasetName, and used it for the subsequent analyses and benchmark construction.

\looseness=-1
\datasetName~consists of prompts, responses, and vote counts.
Across the 908 prompts, each prompt has approximately 96 responses and 172 votes, on average.
The total number of prompt--response pairs is 82,536.
This is approximately seven times larger than that of Oogiri-GO \cite{Zhong_2024_CVPR} and, to the best of our knowledge, is the largest Japanese Oogiri dataset to date.\footnote{Compared with 11,842 Japanese Oogiri instances in a text-to-text setting.}
Moreover, although Oogiri-GO averages approximately eight responses per prompt, our dataset offers approximately 96 responses, yielding a far more diverse candidate set per prompt.
This breadth enables raters to select responses that are genuinely funny rather than merely ``less bad'' within a limited pool.
Dataset statistics are presented in Table \ref{tab:dataset_statistics}.

\begin{table}[t]
  \centering
  \small
  \begin{tabular}{lr}
    \toprule
    \bf{Statistic} & \bf{Value} \\
    \midrule
    Prompts                              & 908    \\ 
    Responses per prompt (avg.)          & 95.9   \\ 
    Votes per prompt (avg.)              & 171.6  \\ 
    Votes per response (avg.)            & 1.8    \\ 
    Votes per top-1 response (avg.)      & 10.3   \\ 
    Prompt length in characters (avg.)   & 20.4   \\ 
    Response length in characters (avg.) & 16.4   \\ 
    \bottomrule
  \end{tabular}
  \caption{Summary statistics of \datasetName.}
  \label{tab:dataset_statistics}
\end{table}

\section{Linguistic Feature Analysis}
\label{sec:linguistic_feature_analysis}
\looseness=-1
We address the first challenge mentioned in \S\ref{sec:introduction}: elucidating the components that constitute a ``funny response.''
``Funniness'' is subjective and complex; for example, it involves expectation violations and relatability. However, a generally accepted quantitative metric remains lacking.
Accordingly, our analysis aims to explain and analyze why an Oogiri response is funny based on a variety of quantitative linguistic features.
Through this analysis, we seek to identify objective and quantitative indicators for understanding humor and to pave the way for improving the ability of LLMs to understand humor.

\subsection{Dataset for Analysis}
\label{subsec:dataset_for_analysis}
We quantitatively examined the linguistic features that constitute ``humor,'' using \datasetName~as the foundation.
Although the dataset links an average of 96 responses to each prompt, we did not use all responses for the analysis.
This is because many responses have zero votes, creating a pronounced imbalance between high-rated responses with many votes and low-rated responses with no votes, which makes the analysis challenging.

Accordingly, we first narrowed down the responses under analysis and balanced the high- and low-rated responses.
Specifically, for each prompt, we defined the top three responses by vote count as ``high-rated responses'' and the bottom three as ``low-rated responses.''
On average, high-rated responses received approximately 8.5 votes, whereas all low-rated responses had zero votes.
Given this low-rated nature, we considered them as reasonable representatives of ``unfunny responses.''
This yielded 5,448 responses for the analysis, with 908 prompts $\times$ 6 responses.


\subsection{Analysis Methodology}
\label{subsec:analysis_methodology}
We examined the relationships between linguistic features and response humor.
Specifically, for each response, we quantitatively measured a range of linguistic features and analyzed the relationship of these feature values to response humor (i.e., differences between the high- and low-rated groups).
We defined and quantified various aspects of linguistic features by borrowing ideas from theories of humor, such as incongruity theory~\cite{Morreall2024PhilosophyHumor}.
These include basic linguistic features, such as sentence length, as well as higher-order features, such as resolution of incongruity (see details in \S\ref{subsec:linguistic_features}).
We considered that, when a feature exhibits a significantly higher or lower value in high-rated responses, it may constitute a component of humor.

We reported these relationships using an independent two-sample Student’s t-test (two-sided, assuming equal variances)~\cite{fisher1925statistical} and Cohen's d~\cite{cohen1988spa}.
The t-test assesses whether there is a statistically significant difference between two group means. 
Because the t-tests are sensitive to large sample sizes, we also reported Cohen's d, an effect-size measure.
Cohen's d is the difference between the two group means divided by a pooled standard deviation and is used to evaluate the magnitude of the effect.
Larger values indicate more substantively meaningful group differences.
The formula for Cohen's d is as follows:
\[
d = \frac{\bar{X}_1 - \bar{X}_2}{s_p}, \quad
s_p = \sqrt{\frac{(n_1-1)s_1^2 + (n_2-1)s_2^2}{n_1 + n_2 - 2}},
\]
where $\bar{X}$, $s$, and $n$ are the mean, standard deviation, and sample size for each group, and $s_p$ is the pooled standard deviation.
The conventional benchmarks interpreted $d=0.2$, $0.5$, and $0.8$ as small, medium, and large effects, respectively.

\subsection{Linguistic Features}
\label{subsec:linguistic_features}
To capture humor from multiple perspectives, we defined four groups of features listed in Table~\ref{tab:linguistic_features_analysis} and measured them quantitatively.
Inspired by the theories of humor~\cite{Morreall2024PhilosophyHumor} and prior research on humor and other creative domains~\cite{Zhong_2024_CVPR,murakami-etal-2025-adparaphrase-v2}, we selected these features as plausible constituents of humor.

\paragraph{Basic Linguistic Features}
We defined basic linguistic features that 
comprise (i) response-independent measures and (ii) prompt--response relative measures.
The former is based solely on the response, whereas the latter is based on the relationship between the prompt and response.
The response-independent measures include sentence length based on character count, number of unique characters, ratios of character types (e.g., \textit{hiragana} and \textit{katakana} in Japanese), and POS ratios (e.g., nouns, verbs, and symbol marks).
We used a Japanese morphological analyzer, MeCab~\cite{kudo-etal-2004-applying}, to perform tokenization and POS tagging.
The prompt--response relative measures include length ratios of prompt--response pairs based on character count, lexical novelty ratios, and relative change in character-type ratios.
We defined the lexical novelty ratio as the proportion of words in the response that do not appear in the prompt and the relative change in character-type ratios as the difference in the ratios of character types between the prompt and response.

\paragraph{Semantic Distance and Textual Entailment}
\looseness=-1
Inspired by incongruity theory~\cite{mcdonald2013philosophy}, we introduced semantic features that capture how a response deviates from the expectations set by the prompt.
Incongruity theory states that \textit{humor arises when expectations are violated}~\cite{Morreall2024PhilosophyHumor}.
In a prompt--response setting, this corresponds to semantic divergence or explicit contradiction between the two texts.
To capture this relationship, we used two signals: (i) semantic distance and (ii) textual entailment.
Semantic distance is measured as one minus the cosine similarity between the prompt and response embeddings. 
Textual entailment is measured using natural language inference (NLI) probabilities, namely entailment, neutral, and contradiction, predicted using an NLI model.
We used the \texttt{text-embedding-3-large}~\cite{OpenAITextEmbedding} to obtain the text embeddings and the \texttt{mDeBERTa-v3-base}~\cite{he2021deberta} fine-tuned on the XNLI~\cite{conneau-etal-2018-xnli} and \texttt{multilingual-NLI-26lang-2mil7} datasets~\cite{laurer_less_2022} to obtain the NLI probabilities.\footnote{\url{https://huggingface.co/MoritzLaurer/mDeBERTa-v3-base-xnli-multilingual-nli-2mil7}}
We assumed that higher semantic distance or explicit contradiction indicates higher unexpectedness.
We then quantitatively tested whether leveraging contradictions increases the degree of humor.

\paragraph{Surprisal and Pointwise Mutual Information}
In addition to the aforementioned features grounded in incongruity theory, we introduced two metrics by borrowing ideas from information theory: surprisal~\cite{Shannon1948-mtc} and normalized pointwise mutual information (nPMI)~\cite{fano1961transmission}.
Surprisal is the length-normalized negative log-probability under a language model; higher values indicate less predictable responses.
nPMI quantifies the association between a prompt and its response; lower values imply co-occurrence that is close to chance.
Both metrics also capture deviation from expectation in incongruity theory: surprisal reflects unpredictability of the prompt--response pair or response text itself, whereas nPMI captures unexpectedness in the prompt--response relationship.
We computed these using GPT-2.\footnote{\url{https://huggingface.co/rinna/japanese-gpt2-medium}}

\paragraph{LLM-Scored Higher-Order Features}
We used an LLM to measure eight higher-order linguistic features.
By ``higher-order,'' we mean features that extend beyond surface cues (e.g., length) and probabilistic or embedding-based signals (e.g., surprisal).
Therefore, we used an LLM to score each prompt--response pair on a 1--5 scale across the eight aspects listed in Table \ref{tab:linguistic_features_analysis}.
These aspects include the following:
(1) \textit{Ambiguity exploitation}: The use of lexical or structural ambiguity,  
(2) \textit{Associative distance}: A moderate and natural conceptual leap,  
(3) \textit{Benign violation}, grounded in benign violation theory~\cite{McGraw2010-kg}, with deviations framed as harmless and acceptable, 
(4) \textit{Coherence}: Strong discourse-level connectedness,  
(5) \textit{Expectedness}: The ease of predicting the response, 
(6) \textit{Incongruity resolution}, grounded in incongruity-resolution theory~\cite{ritchie2009variants}; the natural resolution of an initial mismatch by a coherent reinterpretation, 
(7) \textit{Metaphor use}: The presence of metaphorical expression in the response, 
(8) \textit{Perspective shift}: A meaningful change in viewpoint or framing that enables a punchline.
In all cases, higher scores indicate more of the stated property.
We defined clear evaluation criteria for each aspect and incorporated them into the prompt.\footnote{The full prompt is provided in Appendix~\ref{appendix:prompt_examples}.}
Because of API cost considerations, we sampled 2,000 prompt--response pairs, where 1,000 pairs were randomly selected from high- and low-rated groups, and conducted batched evaluations for each pair using GPT-5.


\subsection{Analysis Results}
\label{subsec:analysis_results}
\begin{table}[t]
\small
\centering
\begin{tabular}{@{ }l@{\hspace{-1em}}r@{\hspace{0.1em}}r@{\hspace{0.8em}}r@{ }}
\toprule
  \multirow{2}{*}{\bb{Feature names}} &
  \multicolumn{2}{c}{\bb{Feature values}} &
  \multicolumn{1}{c}{\bb{Cohe-}} \\ \cmidrule(lr{0.75em}){2-3}
  & \multicolumn{1}{c}{\bb{High}} & \multicolumn{1}{c}{\bb{Low}} & \multicolumn{1}{c}{\bb{n's d}} \\ 
  \midrule
\bb{Basic Features}                           &        &        &            \\
\ \ \textit{\underline{Response-independent}} &        &        &            \\
\ \ \ length$^{*\dagger}$                     & 14.12  & 16.40  & \bb{-0.28} \\
\ \ \ unique chars$^{*\dagger}$               & 13.24  & 15.32  & \bb{-0.30} \\
\ \ \ hiragana ratio$^{*}$                    & 0.46   & 0.44   & 0.11       \\
\ \ \ katakana ratio$^{*\dagger}$             & 0.14   & 0.16   & -0.11      \\
\ \ \ noun ratio$^{*}$                        & 0.42   & 0.45   & -0.13      \\
\ \ \ verb ratio$^{*}$                        & 0.16   & 0.14   & 0.10       \\
\ \ \ symbol ratio$^{*\dagger}$               & 1.91   & 2.24   & -0.07      \\
\ \ \textit{\underline{Prompt-response}}      &        &        &            \\
\ \ \ length ratio$^{*\dagger}$               & 0.76   & 0.90   & \bb{-0.27} \\
\ \ \ lexical novelty$^{*}$                   & 0.80   & 0.93   & \bb{-0.21} \\
\ \ \ hiragana changes$^{*}$                  & -0.04  & -0.06  & 0.10       \\
\ \ \ katakana changes$^{*}$                  & 0.02   & 0.05   & -0.10      \\
\bb{Semantic / NLI}                           &        &        &            \\
\ \ semantic distance$^{*}$                 & 0.73   & 0.72   & 0.16       \\
\ \ contradiction$^{*}$                       & 0.28   & 0.27   & 0.06       \\
\ \ entailment$^{*}$                          & 0.17   & 0.14   & 0.18       \\
\ \ neutral$^{*}$                             & 0.55   & 0.59   & -0.15      \\
\bb{Surprisal / PMI}                          &        &        &            \\
\ \ nPMI$^{*}$                                & 0.12   & 0.14   & -0.14      \\
\ \ surprisal$_{\textrm{response-independent}}^{*}$  & 5.17   & 5.08   & 0.08       \\
\ \ surprisal$_{\textrm{prompt-response}}^{*}$        & 4.66   & 4.51   & 0.13       \\
\bb{LLM-Scored Features}                      &        &        &            \\
\ \ ambiguity exploitation$^{*\dagger}$       & 2.10   & 1.61   & \bb{0.42}  \\
\ \ associative distance$^{*\dagger}$         & 4.38   & 3.90   & \bb{0.33}  \\ 
\ \ benign violation$^{*\dagger}$             & 4.73   & 4.49   & \bb{0.27}  \\
\ \ coherence$^{*}$                           & 4.11   & 3.95   & 0.15       \\
\ \ expectedness                              & 2.68   & 2.78   & -0.08      \\
\ \ incongruity resolution$^{*\dagger}$       & 3.71   & 3.35   & \bb{0.36}  \\
\ \ metaphor use$^{*\dagger}$                 & 1.54   & 1.31   & \bb{0.24}  \\
\ \ perspective shift$^{*\dagger}$            & 2.40   & 1.87   & \bb{0.50}  \\ \bottomrule
\end{tabular}
\caption{Comparison of linguistic features between high- and low-rated responses. 
$*$ indicates statistical significance ($p < 0.05$). Bold values in the Cohen's d indicate a small or medium effect size ($|d| \geqq 0.2$). $^\dagger$ indicates features that are employed in the benchmark experiments~(\S\ref{sec:benchmark_experiments}). }
\label{tab:linguistic_features_analysis}
\end{table}

We report on the relationships between each linguistic feature and response humor.
Table \ref{tab:linguistic_features_analysis} presents the mean of each feature for the high- and low-rated groups, p-value of the t-test, and Cohen’s d.
Our analysis yielded the following findings:

\paragraph{High-Rated Responses Tend to be Shorter}
\looseness=-1
Length-related features such as the length and prompt--response length ratios were significantly lower in the high-rated group than in the low-rated group, with small effect sizes.
This suggests that brevity contributes to humor.

\paragraph{Appropriate Vocabulary Diversity is Beneficial}
\looseness=-1
Interestingly, the high-rated group showed significantly lower values for the unique character count (unique chars) and the rate at which vocabulary that is not in the prompt appears in the response (lexical novelty), with small effect sizes.
This indicates that, relative to the low-rated group, high-rated responses had a lower tendency to use new vocabulary and may benefit from selecting appropriate words without straying far from the topic of the prompt.

\paragraph{Higher-Order Linguistic Features are Effective}
\looseness=-1
Ambiguity exploitation, associative distance, benign violation, incongruity resolution, metaphor use, and perspective shift were significantly higher in the high-rated group, with small-to-medium effect sizes. 
Among these, perspective shift and ambiguity showed relatively larger effects, indicating particular importance for humor.
Incongruity resolution, grounded in incongruity-resolution theory~\cite{ritchie2009variants}, also showed a relatively large effect size, suggesting its contribution to humor.

\paragraph{Other Features Have Limited Impact}
\looseness=-1
Semantic distance, textual entailment, surprisal, nPMI, and other linguistic features (e.g., POS ratio) showed statistically significant differences, but the effect sizes were below small, suggesting limited contributions to humor.
Notably, textual entailment and surprisal captured similar aspects to coherence and expectedness in higher-order linguistic features, but their effect sizes were below small, consistently suggesting their limited role in constituting humor.


\section{Oogiri Understanding Benchmark \label{sec:benchmark_experiments}}
We propose a novel benchmark, \benchmarkName. 
The aim of this benchmark is to measure the ability of an LLM to understand and judge ``humor'' in Oogiri from different perspectives.
Specifically, we propose five tasks that can be broadly grouped into two categories: four relative-judgment tasks using multiple-choice question answering (MCQA) and one absolute-judgment task using binary classification. 
Standardized prompt templates and strict evaluation criteria were used to ensure reproducibility and comparability. 
In the experiments, we tested the insights from our analysis results in \S\ref{sec:linguistic_feature_analysis} and reflected the multiple linguistic features into prompt templates, seeking the performance gains of LLMs (\S\ref{subsec:benchmark_experiments_details}).
Our goal was to clarify the current state of LLM humor understanding and outline a path for further improvement.

\subsection{Task Design}
\label{subsec:task_design}
\looseness=-1

\paragraph{Relative Judgment Tasks}
In the MCQA setting, the model selects the most humorous response to a given prompt from several candidate responses.
We defined four types of tasks: two binary-choice tasks, a three-choice task, and a four-choice task.
In all tasks, the high-rated response for each prompt served as the positive example, and the negatives were constructed differently for each task.
For the two binary-choice tasks, we constructed negatives in two ways: (i) we paired the positive with one low-rated response from the same prompt (Binary$_{\textsf{same}}$) and (ii) we paired the positive with one high-rated response for a different prompt (Binary$_{\textsf{diff}}$).
The latter evaluates whether the model can judge funniness as a response to the given prompt, rather than merely ranking responses within the same prompt, following \citet{hessel-etal-2023-androids}.
For the three- and four-choice tasks, we used one low-rated same-prompt response and one or two high-rated different-prompt responses as negatives, respectively.

\paragraph{Absolute Judgment Task}
In the binary classification setting, the model decides whether a response to a prompt is ``funny'' or ``not funny.''
For each prompt, we used the high-rated response as the positive and the low-rated response as the negative, measuring the ability of the model to evaluate funniness in absolute terms.
Figure~\ref{fig:absolute-prompt-example} shows an example of the absolute-judgment prompt.

\input{figures/absolute-prompt-example}

\subsection{Dataset Construction}
\label{subsec:benchmark_dataset_construction}
\benchmarkName~is built on \datasetName.
For the MCQA setting, we sampled 100 prompts per task from \datasetName, and selected positives and negatives according to each task design, yielding 400 items across the four tasks.
For binary classification, we sampled 100 prompts from \datasetName, pairing one high-rated response and one low-rated response per prompt for 200 items.
In total, \benchmarkName~comprised 600 items.\footnote{To prevent data contamination, we sampled different data points from the analysis dataset in \S\ref{sec:linguistic_feature_analysis}.}

\subsection{Benchmark Experiments}
\label{subsec:benchmark_experiments_details}

\subsubsection{Experimental Setup}
\looseness=-1
We evaluated a range of LLMs listed in Table~\ref{tab:benchmark_results_overview}, from proprietary (e.g., GPT-5) to open-source (e.g., DeepSeek-R1), on five tasks in \benchmarkName. 
We report the accuracy as an evaluation metric.
For API-based models, we averaged results over three trials. 
During inference, we set the temperature parameter to zero for all models.

\looseness=-1
We compared two prompting strategies when instructing the LLMs to solve each task. 
(1) a \textit{baseline prompt} that simply instructs the model to select options, as shown in Figure~\ref{fig:absolute-prompt-example}, 
(2) an \textit{insight-augmented prompt} that incorporates features computed from given prompt--response pairs based on the findings of our data analysis.
To keep the prompts concise, we included only a small set of features selected with reference to the observed effect sizes in Table~\ref{tab:linguistic_features_analysis}.
Specifically, we used five basic features: length, unique character count, prompt--response length ratio, symbol ratio, and katakana ratio; and six LLM-scored features: ambiguity exploitation, associative distance, benign violation, incongruity resolution, metaphor use, and perspective shift.
The basic features were precomputed and inserted directly into the prompt.
LLM-scored features followed a two-step procedure: first, for each prompt--response pair, the target LLM computed scores for each aspect (e.g., metaphor use); second, these scores were included as context when instructing the model to select the options for each task.

\looseness=-1
To validate the human performance on this benchmark, we recruited crowdworkers from the crowdsourcing platform\footnote{\url{https://crowdsourcing.yahoo.co.jp/}} and asked them to solve each item using the same baseline prompt that was shown to the LLMs.
Each item was answered by 21 workers, and the final labels were determined by majority vote. 
We included attention checks with unambiguous answers and aggregated the results only for the 21 workers who passed the checks for each item.



\subsubsection{Results and Discussion\label{subsec:benchmark_experiment_results}}
\begin{table*}[t]
\centering
\small
\begin{tabular}{@{ }l@{ }ccc@{\hspace{0.5em}}c@{\hspace{0.5em}}cc@{\hspace{0.8em}}c@{\hspace{0.5em}}c@{ }}
  \toprule
  \multirow{2}{*}{\bf{Models}} & 
  \multirow{2}{*}{\bf{Features}} & 
  \bf{Absolute.} & 
  \multicolumn{4}{c}{\bf{Relative.}} & 
  \bf{Ave.} & 
  \bf{$\Delta$Ave.} \\ \cmidrule(lr{0.75em}){3-3} \cmidrule(lr{0.75em}){4-7}
  &
  &
  \bf{Binary$_{\textsf{class}}$} & 
  \bf{Binary$_{\textsf{diff}}$} & 
  \bf{Binary$_{\textsf{same}}$} & 
  \bf{Triple} & 
  \bf{Quad} & 
  \bb{Accuracy} & 
  \bb{Accuracy} \\ \midrule
  \bf{Open LLMs} & & & & & & & & \\
  \ \ gpt-oss-20b                      & --         & 50.5      & 64.0       & 45.0      & 33.0      & 37.0       & 45.9      & --        \\
  \ \ gpt-oss-20b                      & \checkmark & 54.0      & 52.0       & 57.0      & 27.0      & 22.0       & 42.4      & -3.5      \\
  \ \ DeepSeek-R1-14b                  & --         & 48.5      & 56.0       & 43.0      & 31.0      & 28.0       & 41.3      & --        \\
  \ \ DeepSeek-R1-14b                  & \checkmark & 46.0      & 57.0       & 49.0      & 24.0      & 31.0       & 41.4      & +0.1      \\  
  \ \ DeepSeek-R1-14b$_{\textsf{ja}}$  & --         & 52.0      & 61.0       & 42.0      & 38.0      & 30.0       & 44.6      & --        \\
  \ \ DeepSeek-R1-14b$_{\textsf{ja}}$  & \checkmark & 50.0      & 59.0       & 53.0      & 44.0      & 24.0       & 46.0      & +1.4      \\
  \ \ LLM-jp-3.1-13b$_{\textsf{ja}}$   & --         & 47.0      & 80.0       & 45.0      & 39.0      & 38.0       & 49.8      & --        \\
  \ \ LLM-jp-3.1-13b$_{\textsf{ja}}$   & \checkmark & 50.5      & 58.0       & 45.0      & 30.0      & 28.0       & 42.3      & -7.5      \\
  \bb{Proprietary LLMs} & & & & & & & & \\
  \ \ Claude-Opus-4                    & --         & 57.2      & 83.0       & \bb{70.0} & 63.0      & \bb{70.3}  & 68.7      & --        \\
  \ \ Calude-Opus-4                    & \checkmark & 50.8      & 72.7       & 68.0      & 53.0      & 51.3       & 59.2      & -9.5      \\
  \ \ Gemini-2.5-Pro                   & --         & 51.3      & 62.0       & 61.7      & 46.3      & 45.7       & 53.4      & --        \\
  \ \ Gemini-2.5-Pro                   & \checkmark & 50.8      & 58.7       & 66.3      & 51.3      & 47.0       & 54.8      & +1.4      \\
  \ \ GPT-5                            & --         & \bb{61.7} & 89.7       & 65.3      & 62.3      & 59.0       & 67.6      & --        \\
  \ \ GPT-5                            & \checkmark & 60.0      & 93.3       & 69.0      & \bb{69.0} & 62.0       & \bb{70.7} & \bb{+3.1} \\ \midrule
  human                                & --         & 54.5      & \bb{95.0}  & 59.0      & 67.0      & 68.0       & 68.7      &           \\ \bottomrule
\end{tabular}
\caption{Results of benchmark experiments. The best results for each column are \bb{bolded}. ``Ave. Accuracy'' indicates the average accuracy (\%) across five tasks, and ``$\Delta$Ave. Accuracy'' indicates the difference in average accuracy (\%) when using features from our analysis.}
\label{tab:benchmark_results_overview}
\end{table*}

Table \ref{tab:benchmark_results_overview} lists the benchmark results.
We compared two prompting strategies: a baseline prompt and an insight-augmented prompt.

\paragraph{Baseline Prompt}
\looseness=-1
When averaging the accuracy across the five tasks, Claude-Opus-4 performed the best (68.7\%), followed by GPT-5 (67.6\%) and Gemini-2.5-Pro (53.4\%). Open LLMs lagged behind these proprietary LLMs; even the strongest, LLM-jp-3.1-13b$_{\textsf{ja}}$, reached only 49.8\%.
Additionally, with the same instructions as those provided to the LLMs, the 21 crowdworkers achieved 68.7\%, which is comparable to that of Claude-Opus-4.
One possible reason that the human performance was relatively low compared with our expectations is the demographic mismatch between crowdworkers and users of the Oogiri platform.\footnote{Because neither the crowdsourcing service nor the Oogiri platform discloses detailed user attributes, we could not perform a precise comparison; however, some differences in user populations are plausible.}
Humor is subjective, and differences in age and interests can yield different judgments of funniness.
Future studies will include analyses that account for annotator attributes and evaluations using more diverse raters.

\paragraph{Insight-Augmented Prompt}
With feature incorporation, four models, namely GPT-5, Gemini-2.5-Pro, DeepSeek-R1, and DeepSeek-R1$_{\textsf{ja}}$, improved their average accuracy across the five tasks.
Notably, GPT-5 increased from 67.6\% to 70.7\% (+3.1\%), surpassing both human performance and Claude-Opus-4 in the baseline setting.
This supports the effectiveness of the linguistic features that reflect the components of humor in improving Oogiri understanding.
However, three models, namely Claude-Opus-4, gpt-oss-20b, and LLM-jp-3.1-13b$_{\textsf{ja}}$, degraded. 
One possible factor is differences in the reasoning ability.
Compared with the baseline, the insight-augmented prompt was longer and more complex because of the added features and instructions. 
Stronger reasoners (e.g., GPT-5) could correctly interpret these complex prompts and benefits, whereas weaker models (e.g., LLM-jp-3.1-13b$_{\textsf{ja}}$) tended to misinterpret them and over-rely on feature magnitudes. 
For example, given the insight that funnier responses tend to be shorter, weaker models over-selected very short responses. 
This suggests that when reasoning is limited, instructing models to consider features can introduce overfitting problems and reduce performance.

\subsubsection{Analysis}

\paragraph{Effectiveness of Continued Pretraining on Japanese Corpus}
\looseness=-1
We compared the two models in Table \ref{tab:benchmark_results_overview}, namely DeepSeek-R1 and DeepSeek-R1$_{\textsf{ja}}$, which share the same architecture and parameter count; the only difference is the pretraining data.
DeepSeek-R1$_{\textsf{ja}}$ continues pretraining DeepSeek-R1 on a Japanese corpus.\footnote{\url{https://huggingface.co/cyberagent/DeepSeek-R1-Distill-Qwen-14B-Japanese}}
DeepSeek-R1$_{\textsf{ja}}$ improved the average accuracy across the five tasks from 41.3\% to 44.6\% in the baseline setting (+3.3 points) and from 41.4\% to 46.0\% in the insight-augmented setting (+4.6 points). 
As our benchmark is based on Japanese Oogiri, these results suggest that continued pretraining on a Japanese corpus is effective in improving Oogiri understanding.
Although prior work has shown benefits for Japanese cultural and knowledge understanding~\cite{tsutsumi-jinnai-2025-large}, our findings indicate that such continued pretraining aids in the more advanced language understanding required for Japanese Oogiri.

\begin{table}[t]
\small
\centering
\begin{tabular}{@{ }l@{\hspace{0.3em}}c@{\hspace{0.5em}}c@{\hspace{0.5em}}cc@{ }}
\toprule
  \multirow{2}{*}{\bf{Models}} & 
  \multicolumn{2}{c}{\bf{Features}} & 
  \bf{Ave.} & 
  \bf{$\Delta$Ave.} \\ 
  &
  \bf{Basic.} & 
  \bf{LLM-scored.} & 
  \bf{Acc.} &  
  \bf{Acc.} \\ 
  \midrule
Gemini-2.5-Pro & --         & --         &  53.4       & --          \\
Gemini-2.5-Pro & \checkmark & --         &  \bb{57.1}  & \bb{+3.7}   \\
Gemini-2.5-Pro & --         & \checkmark &  54.5       & +1.1       \\
Gemini-2.5-Pro & \checkmark & \checkmark &  54.8       & +1.4       \\ \midrule
GPT-5          & --         & --         &  67.6       & --          \\
GPT-5          & \checkmark & --         &  69.8       & +2.2       \\ 
GPT-5          & --         & \checkmark &  68.0       & +0.4       \\
GPT-5          & \checkmark & \checkmark &  \bb{70.7}  & \bb{+3.1}   \\ \bottomrule
\end{tabular}
\caption{Ablation study on feature types. 
``$\Delta$Ave. Acc.'' represents the difference in average accuracy compared to the model without any features.
The best accuracy for each model is \bb{bolded}.}\label{tab:ablation_study_feature_types}
\end{table}

\paragraph{Ablation Study of Feature Groups}
Table \ref{tab:ablation_study_feature_types} presents the average accuracy over the five tasks for GPT-5 and Gemini-2.5-Pro under four settings: introducing only basic linguistic features, introducing only LLM-scored higher-order features, introducing both, and using the baseline with no features.
In all cases, incorporating features into a prompt improved the average accuracy over the baseline.
For GPT-5, using both feature groups yielded the best results.
For Gemini-2.5-Pro, introducing only basic linguistic features (e.g., length and character-type ratios) performed the best.
Notably, when introducing only basic linguistic features, both Gemini-2.5-Pro and GPT-5 improved more than when introducing higher-order features alone (e.g., +3.7 and +2.2 points, respectively).
Response length was already identified in our analysis as a constituent component of humor, and the benchmark results empirically confirm that such simple heuristics can be effective criteria for evaluating funniness.
These findings suggest that exploring a broad range of linguistic features is a promising direction for enhancing the humor understanding of LLMs further.

\begin{table}[t]
\centering
\small
\begin{tabular}{@{ }lccc@{}} \toprule
\textbf{Models} &
\textbf{Features} &
\textbf{Uncertain} &
\bf{Ave. Accuracy} \\ \midrule
GPT-5 & --          & --         & 67.6 \\
GPT-5 & \checkmark  & --         & 68.9 \\
GPT-5 & \checkmark  & \checkmark & 70.7 \\
\bottomrule
\end{tabular}
\caption{Ablation study on instruction styles. ``Uncertain'' indicates whether the uncertain instruction style is used. ``Ave. Accuracy'' indicates the average accuracy (\%) across five tasks.}
\label{tab:ablation_instruction_styles}
\end{table}

\paragraph{Effect of Instruction Style for Feature Use}
We also examined the influence of instruction style on  performance when incorporating features into prompts, that is, how we should tell the model to use the features.
We considered two styles: (1) instructing the model to use the features when judging funniness, and (2) instructing the model to consult the features \textit{only when uncertain}.
In our preliminary experiments, we first attempted style (1) and observed an over-reliance on feature magnitudes, which motivated the proposal of style (2).
Table \ref{tab:ablation_instruction_styles} shows the average accuracy of GPT-5 over the five tasks for the no-feature baseline and the two instruction styles. 
Here, the ``Uncertain'' column corresponds to style (2).
In both styles, incorporating features improved over the baseline; notably, style (2) yielded the highest performance, improving the average accuracy by 3.1 points over the baseline.
This indicates that asking the model to consider features only when uncertain helps to prevent over-dependence on feature magnitudes and enables more appropriate use of the features.
The results highlight instruction design as an important lever for improving the humor understanding of LLMs, and the value of exploring more effective instruction styles in future studies.

\section{Conclusion}
\label{sec:conclusion}
We presented a systematic study of humor on \datasetName, and introduced \benchmarkName, a benchmark covering relative and absolute judgments.
Our analysis showed that multiple linguistic features, such as length and ambiguity, correlated with high-rated responses.
In the benchmark experiments, we showed that incorporating these features into prompts improves the model performance.
Furthermore, we demonstrated that continued pretraining on a Japanese corpus further boosts accuracy and instructing models to consider features only when uncertain mitigates over-reliance on heuristics.
Future work will include exploring other effective linguistic features and refining prompt design, scaling human evaluations with annotator attributes, and extending the method to other languages and multimodal settings.


\section{Ethics Statements} 
\label{sec:ethics_statements}
\paragraph{Data Collection and Licensing}
\looseness=-1
\datasetName~was constructed by collecting data from the public Japanese Oogiri competition platform, Oogiri Sogo.
We confirm that the site explicitly permits web crawling, ensuring the legitimacy of the data collection process in \S\ref{sec:dataset_construction}.
To promote transparency and facilitate further research, \datasetName~and \benchmarkName~will be made available under the CC BY-NC-SA 4.0 license.

\paragraph{Human Evaluation on \benchmarkName}
\looseness=-1
We recruited crowdworkers for human baseline evaluation in \S\ref{sec:benchmark_experiments}. 
We used Yahoo! Crowdsourcing as the crowdsourcing platform.
In accordance with the platform's regulations, the compensation was set at 10 yen per 20 tasks.
Workers were informed that the annotated results would be used for research purposes.
In addition, we acknowledge that a potential demographic mismatch between the crowdworkers and Oogiri-platform users exists as discussed in \S\ref{subsec:benchmark_experiment_results}, suggesting that a further analysis accounting for annotator attributes is necessary to improve the evaluation reliability.

\section{Limitations}
\label{sec:limitations} 
\paragraph{Limited to Japanese Oogiri}
Our analysis and benchmark are based on Japanese Oogiri data. Some humor depends on culture-specific knowledge (e.g., a response such as ``Mount Fuji'' may be funny to Japanese users because it evokes familiar shared knowledge), and similar effects may not hold in other languages or cultural contexts.
Moreover, our feature analysis included Japanese-specific elements (e.g., character-type ratios), which may not be directly transferred. 
Future work should include collecting and analyzing Oogiri-like data in other languages and cultures to better understand the cross-lingual and cross-cultural variations in humor.

\paragraph{Benchmark Scope Limited to Oogiri Understanding} 
We proposed a benchmark focused on understanding ``funniness'' in Oogiri: four MCQA subtasks and one binary classification task.
However, humor understanding is related to other capabilities such as generation and explanation \cite{Loakman2025INLG}.
Although these are beyond the scope of this study, extending the benchmark to evaluate generation and explanation is an important direction for future research.

\paragraph{Focus on Unimodal Settings}
As discussed in Related Work (\S\ref{sec:related_work}), Oogiri can be framed as text-to-text, image-to-text, or image\&text-to-text \cite{Zhong_2024_CVPR}.
We focused on the text-to-text approach for two reasons: (1) as a first step toward measuring LLM humor understanding, a unimodal text-only setup reduces complexity relative to multimodal settings, and (2) text-to-text Oogiri data are more abundant on the web, facilitating robust dataset construction and generalizable analysis.
An important next step is to extend the dataset to multimodal variants and study humor understanding involving visual information.


\section{Bibliographical References}
\bibliographystyle{lrec2026-natbib}
\bibliography{references}


\appendix

\section{Prompt Example\label{appendix:prompt_examples}}
Figure~\ref{fig:prompt-for-llm-features} presents a prompt template used to instruct LLMs to evaluate the higher-order linguistic features for given prompt--response pairs.
\begin{figure*}[t]
  \centering
  \begingroup
  \scriptsize
  \setlength{\baselineskip}{0.92\baselineskip}
  \begin{tcolorbox}[
    colback=black!7,
    colframe=black!50,
    boxrule=0.4pt,
    arc=2pt,
    left=1mm, right=1mm, top=0.6mm, bottom=0.6mm,
    width=\linewidth
  ]
    {\ttfamily
    You receive a prompt and a response, and your task is to evaluate how appropriate the response is for the prompt. Please evaluate the characteristics of the response to the prompt for the following ``Oogiri'' scenario on a scale of 1--5.\\[2pt]
    Prompt: \textbf{\{prompt\}}\\
    Response: \textbf{\{response\}}\\[2pt]
    Evaluate based on the following criteria and respond in JSON format.\\
    In doing so, please explain the reasoning for the scores.\\[2pt]

    1) ambiguity\_exploitation (1--5): Use of Ambiguity\\
    1: The response does not exploit ambiguity.\\
    3: The response is somewhat ambiguous.\\
    5: The response effectively exploits ambiguity.\\[2pt]

    2) associative\_distance (1--5): Appropriateness of Association\\
    1: The association between prompt and response is direct OR requires 5 or more associative leaps.\\
    3: The association is reached in 1 step OR requires 4 steps with somewhat unnatural association.\\
    5: The association is naturally reached in 2--3 steps.\\[2pt]

    3) benign\_violation (1--5): Degree of Harmless Violation\\
    1: The response deviates from the prompt and is extremely harmful/offensive.\\
    3: The response deviates from the prompt and is somewhat harmful/offensive.\\
    5: The response deviates from the prompt but is harmless.\\[2pt]

    4) coherence (1--5): Logical Coherence between prompt and Response\\
    1: The prompt and response are not logically connected.\\
    3: The prompt and response are somewhat logically connected.\\
    5: The prompt and response are perfectly logically connected.\\[2pt]

    5) expectedness (1--5): Predictability of the Response\\
    1: The response is completely unexpected and surprising relative to the prompt.\\
    3: The response is somewhat unexpected or surprising relative to the prompt.\\
    5: The response is very predictable or obvious relative to the prompt.\\[2pt]

    6) incongruity\_resolution (1--5): Degree of Resolution of Incongruity\\
    1: The incongruity between the prompt and response is not resolved at all.\\
    3: The incongruity between the prompt and response is somewhat resolved.\\
    5: The incongruity between the prompt and response is naturally resolved.\\[2pt]

    7) metaphor\_use (1--5): Appropriateness of Metaphor Use\\
    1: The response does not use metaphor regarding the prompt.\\
    3: The response somewhat uses metaphor regarding the prompt.\\
    5: The response uses metaphor regarding the prompt.\\[2pt]

    8) perspective\_shift (1--5): Shift in Perspective\\
    1: The response shows no shift in perspective regarding the prompt.\\
    3: The response shows a partial shift in perspective regarding the prompt.\\
    5: The response shows a clear shift in perspective regarding the prompt.\\[2pt]

    Output Requirements:\\
    - All scores must be integers (1--5).\\
    - In the reasoning field, summarize the concise basis for each score in 1--3 sentences.\\
    - Return in JSON format.\\[2pt]

    \{\\
    \quad ``reasoning'': ``Reason for the scores'',\\
    \quad ``ambiguity\_exploitation'': number,\\
    \quad ``associative\_distance'': number,\\
    \quad ``benign\_violation'': number,\\
    \quad ``coherence'': number,\\
    \quad ``expectedness'': number,\\
    \quad ``incongruity\_resolution'': number,\\
    \quad ``metaphor\_use'': number,\\
    \quad ``perspective\_shift'': number\\
    \}\\
    }
  \end{tcolorbox}
  \endgroup
  \caption{Prompt for LLM-based scoring of higher-order linguistic features.}
  \label{fig:prompt-for-llm-features}
\end{figure*}

\end{document}